\documentclass[runningheads]{llncs}

\usepackage{eccv}

\usepackage{eccvabbrv}

\usepackage{graphicx}
\usepackage{booktabs}
\usepackage{siunitx}
\usepackage{amsmath}
\usepackage{outlines}
\usepackage{comment}
\usepackage{afterpage}
\usepackage{import}
\usepackage{wrapfig}
\usepackage{multirow}

\usepackage[accsupp]{axessibility}  
\usepackage{xcolor}
\definecolor{cvprblue}{RGB}{0,113,187}
\definecolor{lightblue}{RGB}{234,249,255}
\definecolor{grey}{RGB}{235,235,235}
\usepackage{colortbl}
\usepackage{ulem} 
\usepackage{enumitem}
\usepackage[linesnumbered,ruled,vlined]{algorithm2e}
\usepackage{amsmath}
\usepackage{bbding}
\usepackage{tabularx}

\usepackage[breaklinks,colorlinks=true,citecolor=cvprblue,urlcolor=cvprblue]{hyperref}
\usepackage{fontawesome5}

\usepackage{orcidlink}

\everymath=\expandafter{\the\everymath\displaystyle}
\makeatletter\@ifpackageloaded{underscore}{}{\usepackage[strings]{underscore}}\makeatother

\begin{document}

\title{
Efficient Camera Pose Augmentation for View Generalization in Robotic Policy Learning
}

\titlerunning{GenSplat}

\author{Sen Wang\inst{1}, 
Huaiyi Dong\inst{1}, 
Jingyi Tian\inst{1}, 
Jiayi Li\inst{1}, 
Zhuo Yang\inst{1}, \\
Tongtong Cao\inst{2}, 
Anlin Chen\inst{2}, 
Shuang Wu\inst{2}, 
Le Wang\inst{1}, 
Sanping Zhou\inst{1, \text{\Envelope}}}
\authorrunning{S. Wang et al.}

\institute{Xi'an Jiaotong University, Noah's Ark Lab \\
Code is available at: \href{https://github.com/SanMumumu/GenSplat}{\faGithub\ \url{https://github.com/SanMumumu/GenSplat}}
}

\maketitle
{\let\thefootnote\relax\footnotetext{\text{\Envelope} Corresponding author.}}
\vspace{-3ex}

\begin{abstract}




Prevailing 2D-centric visuomotor policies exhibit a pronounced deficiency in novel view generalization, as their reliance on static observations hinders consistent action mapping across unseen views. In response, we introduce GenSplat, a feed-forward 3D Gaussian Splatting framework that facilitates view-generalized policy learning through novel view rendering. GenSplat employs a permutation-equivariant architecture to reconstruct high-fidelity 3D scenes from sparse, uncalibrated inputs in a single forward pass. To ensure structural integrity, we design a 3D-prior distillation strategy that regularizes the 3DGS optimization, preventing the geometric collapse typical of purely photometric supervision. By rendering diverse synthetic views from these stable 3D representations, we systematically augment the observational manifold during training. This augmentation forces the policy to ground its decisions in underlying 3D structures, thereby ensuring robust execution under severe spatial perturbations where baselines severely degrade.

\end{abstract}

\section{Introduction}
\label{sec:intro}

\begin{figure}[t]
    \centering
    \includegraphics[width=\linewidth]{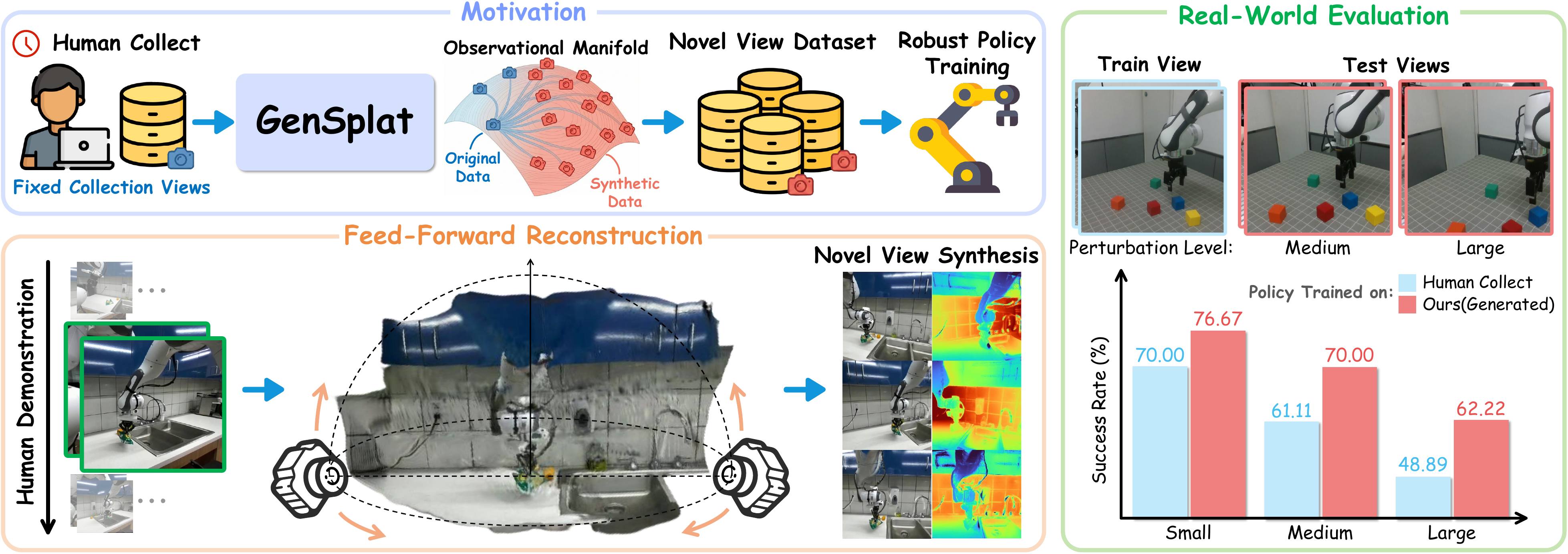}
    \caption{\textbf{The diagram of the proposed GenSplat.} Given expert demonstrations, GenSplat employs a feed-forward 3D reconstruction pipeline to render geometrically consistent novel viewpoints under controlled perturbations, explicitly boosting camera pose diversity. Experiments demonstrate that policies trained on GenSplat-augmented data achieve competitive performance across different perturbation levels, substantially outperforming those trained solely on human-collected demonstrations.
    }
    \vspace{-2ex}
    \label{fig:motivation}
\end{figure}


While recent advances in large-scale imitation learning have demonstrated considerable promise in Embodied AI~\cite{zhao2025cot,black2410pi0,zitkovich2023rt,brohan2022rt,team2024octo,driess2023palm}, their efficacy remains strictly contingent upon expansive, diverse datasets~\cite{o2024open,khazatsky2024droid,walke2023bridgedata,dasari2019robonet}. This data dependency exposes a fundamental vulnerability in the prevailing observation-to-action paradigm: current 2D-centric policies exhibit a pronounced deficiency in novel view generalization. Specifically, because these models inherently rely on perspective-dependent visual representations, even minor extrinsic $SE(3)$ perturbations disrupt the spatial grounding of action-relevant cues~\cite{pumacay2024colosseum,gao2024efficient,abouzeid2025geoaware,pang2025learning}. Mitigating this spatial fragility through the exhaustive re-collection of demonstrations for unseen viewpoints imposes prohibitive marginal costs on teleoperation, thereby hindering scalable real-world deployment~\cite{sadeghi2018sim2real,khazatsky2024droid}. This dilemma naturally motivates a critical inquiry: \textit{How can we augment the observational manifold to induce view generalization without additional human collection?}

To address this, prior efforts have leveraged novel view synthesis (NVS) to artificially expand camera pose diversity, broadly falling into two paradigms: 2D diffusion-based generation~\cite{seo2024genwarp,van2024generative,yang2023movie,zhou2025stable,sargent2024zeronvs} and 3D Gaussian-based reconstruction~\cite{yang2025novel,torne2024robot,li2024robogsim,yu2025real2render2real}.
\textbf{Generative methods} employ diffusion models for plug-and-play viewpoint synthesis, bypassing the need for explicit 3D reconstruction or rigid camera calibration~\cite{yang2023movie,chen2024rovi,tian2024view,shi2025nvspolicy}. Although recent adaptations incorporate 3D-aware conditioning~\cite{sargent2024zeronvs,lu2025movis}, their optimization objectives remain strictly confined to the 2D image plane. Consequently, these generated augmentations lack a unified metric workspace, frequently violating epipolar geometry and multi-view visibility constraints. 
This manifests as severe scale drift, erroneous parallax, and implausible occlusions, which are fatal for precise robotic manipulation. Moreover, underutilizing the multi-camera topology of robotics datasets restricts both geometric fidelity and synthesis controllability.

In contrast, \textbf{Gaussian-based methods} parameterize explicit 3D scenes using anisotropic Gaussian primitives~\cite{kerbl20233d,PapantonakisReduced3dgs}. Rendering from this shared, underlying geometry inherently guarantees strict cross-view physical consistency by explicitly preserving exact occlusion ordering, multi-view parallax, and metric scale, all while enabling real-time synthesis. However, classical 3DGS pipelines suffer from restrictive operational bottlenecks: achieving high-fidelity reconstruction typically necessitates dense multi-view coverage, precise camera pose calibration, and prohibitively expensive per-scene optimization~\cite{yang2025novel,torne2024robot,yu2025real2render2real,duan20244d,li2024robogsim}. Crucially, real-world robotic deployments are fundamentally characterized by extreme viewpoint sparsity, entirely uncalibrated observation setups, and the necessity for instant adaptation across diverse environments. Motivated by this domain gap, we propose a feed-forward reconstruction paradigm capable of instant, high-fidelity scene synthesis from sparse, uncalibrated observations, circumventing the limitations of traditional per-scene optimization.

In this work, we introduce GenSplat, a feed-forward 3D Gaussian Splatting framework designed to expand the observational manifold via novel view synthesis, thereby inducing robust view generalization in visuomotor policies. Circumventing the prohibitively expensive per-scene tuning of classical pipelines, GenSplat leverages recent advances in feed-forward 3DGS~\cite{ye2024no,jiang2025anysplat} to parameterize 3D scenes from sparse, uncalibrated inputs in a single forward pass. This enables efficient, physically grounded multi-view synthesis, systematically expanding the critical dimension of viewpoint diversity required by modern imitation learning architectures~\cite{saxena2025matters,black2410pi0,chi2023diffusion}. However, directly applying naive feed-forward mechanisms to the extreme sparsity of robotic datasets~\cite{o2024open,dasari2019robonet,khazatsky2024droid,walke2023bridgedata} often precipitates severe topological fragmentation and inconsistent occlusions~\cite{kerbl20233d,PapantonakisReduced3dgs}. While such photometric artifacts might be tolerated in pure visual rendering, they severely corrupt the geometric consistency mandatory for reliable visuomotor supervision. To overcome this structural collapse, we propose a 3D-prior distillation objective that extracts geometric knowledge from a 3D foundation model~\cite{wang2025pi} to explicitly regularize the reconstruction landscape. Consequently, GenSplat yields metrically consistent viewpoint augmentations that systematically enhance downstream policy robustness without incurring additional data collection costs.
In summary, our primary contributions are:

(1) We propose GenSplat, a feed-forward 3DGS framework that enables geometrically consistent view augmentation for robotic datasets with sparse and uncalibrated camera observations. The design effectively eliminates the need for sensor calibration, physical viewpoint adjustment, and per-scene optimization.
    
(2) We introduce a 3D-prior distillation strategy that transfers geometric knowledge from a pretrained 3D foundation model to regularize reconstructions, ensuring that synthesized novel views are geometrically consistent and physically reliable for policy learning.

(3) Extensive real-world experiments demonstrate that policies trained with GenSplat-augmented data, achieve strong generalization to unseen camera poses. GenSplat significantly enhances policy performance while demonstrating superior computational efficiency compared to prior methods.

\section{Related Work}
\label{sec:related}
\noindent \textbf{Robot Data Collection Pipelines.} 
The pursuit of generalizable visuomotor policies has catalyzed the curation of large-scale, multi-task robotic datasets~\cite{o2024open,khazatsky2024droid,walke2023bridgedata,dasari2019robonet}. These corpora are typically collected via teleoperation or kinesthetic teaching across diverse environments, pairing observations and natural-language instructions with continuous actions~\cite{iyer2024open,zhang2025kinedex}. While pretraining on such data yields robust initialization~\cite{qu2025spatialvla,zhao2025cot,kim2024openvla,zheng2024tracevla,black2410pi0}, transferring these models to novel physical deployments necessitates target-domain finetuning. Crucially, hardware constraints typically restrict deployment-time collection to a sparse set of fixed extrinsic configurations. Consequently, expanding the observational support to novel viewpoints mandates repeated human demonstrations, imposing an unscalable teleoperation bottleneck~\cite{iyer2024open,patil2024radiance,zhang2025kinedex}. GenSplat fundamentally bypasses this limitation by leveraging feed-forward 3DGS to expand the observational manifold, thereby enriching viewpoint diversity without incurring marginal data collection costs.

\noindent \textbf{3D Gaussian Splatting in Robotics.} 
3DGS has emerged as a practical tool for reconstructing robot workspaces and synthesizing photorealistic observations to support policy learning~\cite{yang2025novel,yu2025real2render2real,huang2025enerverse,li2024robogsim,han2025re,torne2024robot}. 
Several representative systems illustrate this trajectory: Real2Render2Real reconstructs real-world workspaces as Gaussian scenes and exports renderable digital twins for closed-loop policy training and evaluation~\cite{yu2025real2render2real}; RE\(^3\)SIM adopts a real-to-sim pipeline that preserves geometric structure and occlusion relationships to enable realistic manipulation benchmarking~\cite{han2025re}; RoboGSim builds upon Gaussian reconstructions to provide controllable cameras, lighting, and object layouts, facilitating studies on cross-view generalization~\cite{li2024robogsim}; and EnerVerse scales synthetic data generation by composing reusable Gaussian assets with world models to produce diverse, physically plausible scenes~\cite{huang2025enerverse}. 
Recent pipelines further advance novel-view synthesis while retaining metric-scale accuracy in workspace reconstruction~\cite{yang2025novel,torne2024robot,pan2025one}. Collectively, these pioneering works demonstrate that 3DGS enables physically consistent, multi-view rendering and high-fidelity digital twins for robotic applications.
Despite strong geometric fidelity, these pipelines typically require dense, calibrated multi-view capture and costly per-scene optimization, which limits practicality for the sparse and uncalibrated conditions common in robotic manipulation datasets~\cite{o2024open,khazatsky2024droid,walke2023bridgedata,dasari2019robonet}.
GenSplat bypasses these bottlenecks via a feed-forward 3DGS paradigm, directly parameterizing 3D structures from sparse, unposed inputs in a single forward pass. This enables scalable expansion of the observational manifold without rigid calibration or per-scene optimization.

\noindent \textbf{View Generalization Policy Learning.} 
Robust physical deployment necessitates strong $SE(3)$ viewpoint generalization, a challenge as fundamental as task-level generalization~\cite{pang2025learning,yang2023movie,chen2024rovi,tian2024view,shi2025nvspolicy}. Despite the strong in-distribution convergence of modern Vision-Language-Action (VLA) models~\cite{black2410pi0,kim2024openvla,zheng2024tracevla,zhao2025cot,zitkovich2023rt} and Diffusion Policies~\cite{chi2023diffusion}, representations anchored to static camera extrinsics exhibit severe generalization gaps under spatial perturbations. While prior efforts attempt to induce such generalization via architectural inductive biases or algorithmic regularizations~\cite{bai2025learning,seo2023multi,goyal2024rvt2,tian2025pdfactor,wang2025flowram,jiang2025you}, their efficacy remains strictly upper-bounded by the spatial density of the training manifold. Rather than modifying the core policy architecture~\cite{abouzeid2025geoaware,zhang2025grounding,li2025spatial,lin2025evo}, we expand the observational manifold to densify the $SE(3)$ spatial support, inherently inducing view-generalized spatial groundings directly from metrically consistent synthetic observations.

\section{Method}
\label{sec:Methodology}
We detail \textbf{GenSplat}, a feed-forward 3D Gaussian Splatting (3DGS) framework designed to expand the observational manifold via novel view synthesis from sparse, uncalibrated visual demonstrations. By reconstructing metrically consistent 3D workspaces without rigid camera calibration, GenSplat densifies the continuous $SE(3)$ spatial support of the training data, inherently inducing view-generalized action mapping for downstream policies. The overarching architectural pipeline is illustrated in \cref{fig:pipeline}.

\subsection{Problem Formulation}
We consider a robotic imitation learning dataset \( \mathcal{D}=\{(\zeta_i, l_i)\}_{i=1}^{N_\mathcal{D}} \), where each demonstration \(\zeta\) comprises a sequence of observations \(\mathcal{O}=\{I_t\}_{t=1}^T\), paired actions \(\mathcal{A}=\{\boldsymbol{a}_t\}_{t=1}^T\), and a language instruction \(l\). Our goal is to train visuomotor policies that generalize to novel camera views without additional data collection.
To this end, we construct a per-frame, pose-free camera augmentation pipeline, denoted as $\mathcal{G}_\theta$. Given $V$ camera views $\{I_j\}_{j=1}^{V}$ at a time step $t$, $\mathcal{G}_\theta$ represents the 3D scene by a set of Gaussian primitives $\mathcal{S}_t$, predicts the camera parameters for each view $\boldsymbol{p}_t^{v}$, and renders images from sampled target camera poses. The resulting augmented dataset is $\tilde{\mathcal{D}}=\{(\tilde{\zeta}_i, l_i)\}_{i=1}^{N}$, with $\tilde{\zeta}$ replacing each original frame $(I_t, \boldsymbol{a}_t)$ by $(\hat{I}^{\Delta}_{t}, \boldsymbol{a}_t)$, where $\hat{I}^{\Delta}_{t}$ is a novel view rendered from $\mathcal{S}_t$ under a regularly sampled camera transform $\Delta$. The policy $\pi(\boldsymbol{a}_t \mid \mathcal{D}_{\text{train}})$ is trained on the combined dataset $\mathcal{D}_{\text{train}}=\mathcal{D} \cup \tilde{\mathcal{D}}$, inherently inducing view-generalized action mapping.

\subsection{Model Architecture} 
\label{sec3.2 Model Architecture}
To address the reconstruction instability and artifacts caused by improper reference viewpoint selection in traditional methods~\cite{wang2025vggt,keetha2025mapanything,wang2024dust3r}, we design a feed-forward permutation-equivariant architecture. This formulation processes sparse, uncalibrated inputs symmetrically, ensuring robust 3D representation without explicit order dependence. 
To overcome the absence of reliable camera extrinsics in large-scale robotic datasets, the framework integrates geometric distillation from a pre-trained foundation model~\cite{wang2025pi}, providing essential structural regularization for pose-free novel view synthesis.

\textbf{Permutation-Equivariant Architecture.} 
Each image is first processed by DINOv2~\cite{oquab2023dinov2} to generate patch tokens \(\{\boldsymbol{h}_i\}_{i=1}^{V}\in\mathbb{R}^{L\times C}\). To establish a multi-view context without imposing a rigid temporal or spatial order, these tokens are aggregated through alternating intra-frame and global self-attention layers~\cite{wang2025vggt}. This symmetric aggregation mechanism ensures the extracted representations are strictly permutation-equivariant. The tokens are subsequently routed to task-specific transformer decoders. For the \(i\)-th image, a camera pose decoder \(D_\text{c}\) predicts the local-to-global spatial transformation \(\boldsymbol{p}_i = (\boldsymbol{R}_i, \boldsymbol{T}_i) \in \text{SE}(3)\), where \(\boldsymbol{R}_i \in \text{SO}(3)\) denotes the camera rotation and \(\boldsymbol{T}_i \in \mathbb{R}^3\) is the translation vector~\cite{dong2025reloc3r}. Concurrently, a point map decoder \(D_\text{p}\) regresses a dense 3D point map \(\boldsymbol{P}_i\in\mathbb{R}^{H\times W\times3}\) representing the local camera-coordinate geometry, along with a per-pixel confidence map \(\boldsymbol{C}_i\in\mathbb{R}^{H\times W\times1}\).

\begin{figure*}[ht]
    \centering
    \includegraphics[width=1\linewidth]{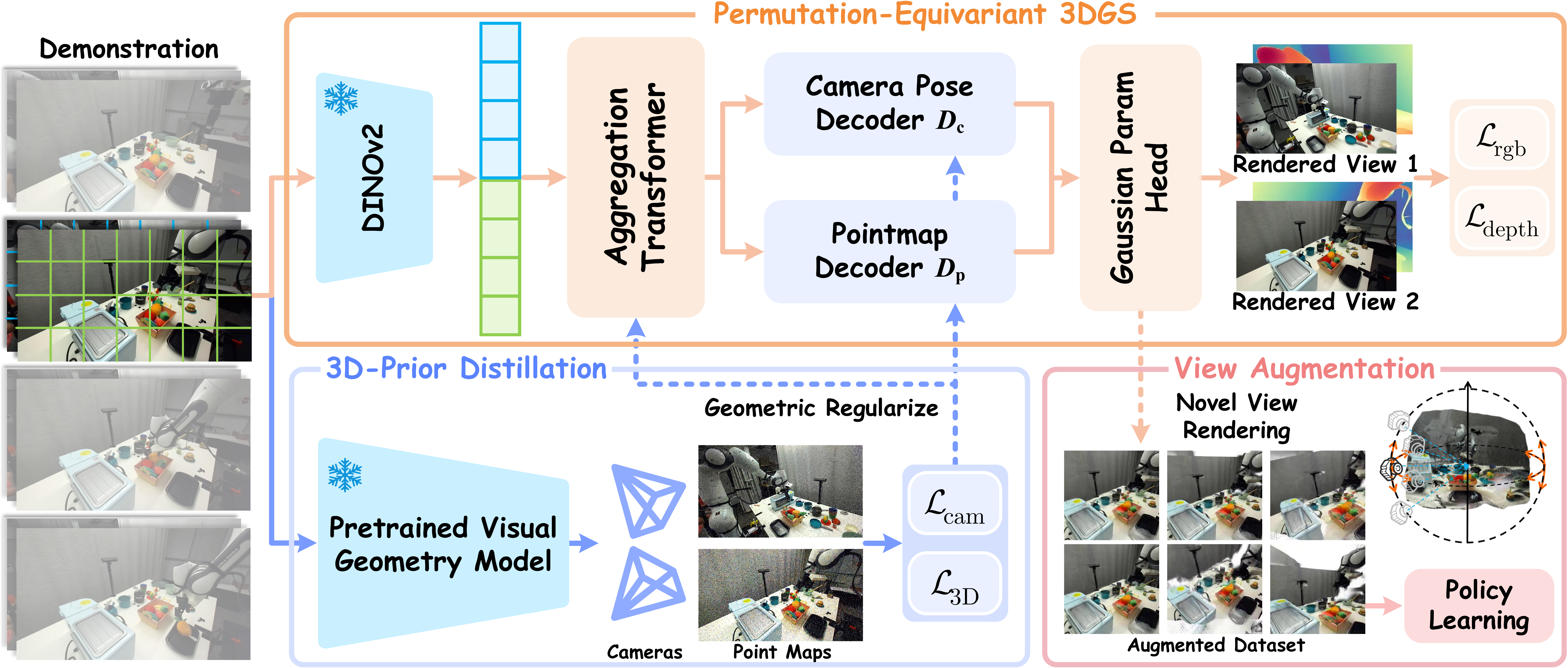}
    \caption{\textbf{Overview of \textit{GenSplat}.} Our feed-forward 3DGS framework reconstructs a 3D scene from sparse, uncalibrated robotic observations. GenSplat employs a permutation-equivariant transformer architecture to predict camera poses, dense point maps, and Gaussian parameters, while leveraging pre-trained visual geometry models for 3D-prior distillation supervision. The reconstructed scenes enable geometrically consistent novel view synthesis to expand the observational manifold, significantly improving the viewpoint generalization of robotic policies.
    }
    \label{fig:pipeline}
    \vspace{-1ex}
\end{figure*}

\begin{figure}[h]
    \centering
    \begin{minipage}[c]{0.48\linewidth}
    As illustrated in~\cref{fig:anysplat_cmp}, we utilize 3D point maps as structural pseudo-labels, unlike AnySplat~\cite{jiang2025anysplat}, which relies on scalar depth maps. Depth-based supervision often fails to maintain rigorous multi-view consistency, rendering the model incapable of reconstructing flat regions and leading to severe structural collapse of the desk. By directly encoding explicit 3D coordinates, point maps offer higher precision and enforce spatial smoothness~\cite{shi2025revisiting,wang2024dust3r}. This effectively prevents geometric collapse, recovering a remarkably accurate and flat desk surface. More visual examples are in Appendix \textcolor{red}{B.2}.
    \end{minipage}
    \hfill
    \begin{minipage}[c]{0.48\linewidth}
        \includegraphics[width=\linewidth]{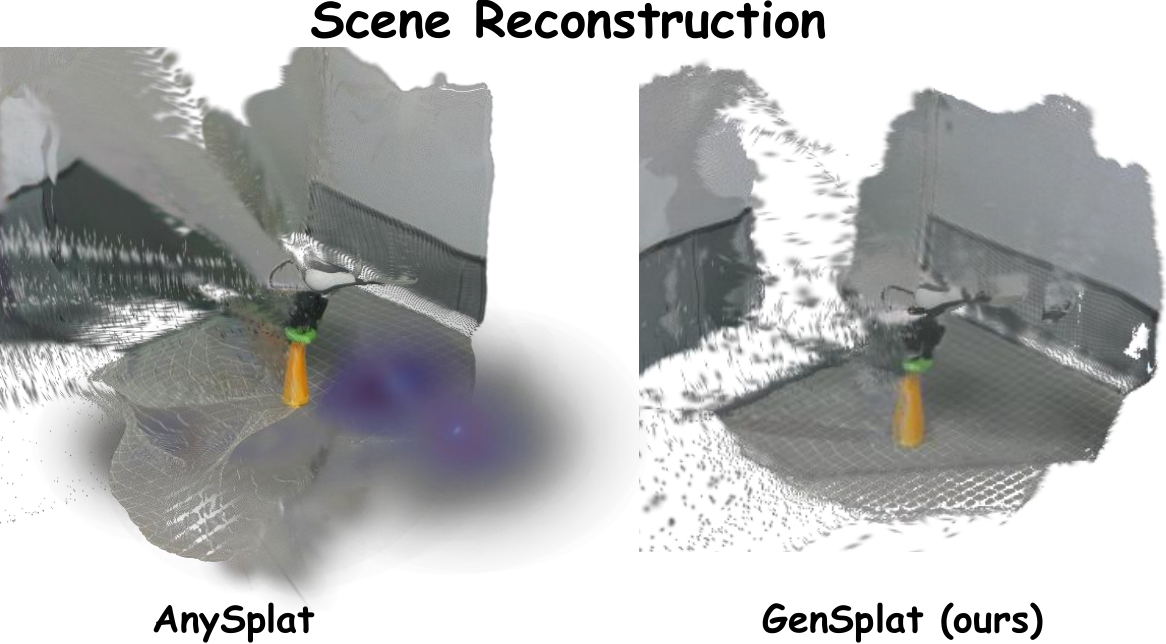}
        \caption{\textbf{Qualitative scene reconstruction comparison.} AnySplat (left) exhibits geometric irregularities on flat surfaces. GenSplat (right) reconstructs accurate, smooth 3D geometry.}
        \label{fig:anysplat_cmp}
    \end{minipage}
    \vspace{-4ex}
\end{figure}

\textbf{Gaussian Parameter Prediction.} 
We explicitly decouple spatial localization from the prediction of rendering attributes. The spatial centers of the Gaussian primitives $\boldsymbol{\mu}_g\in\mathbb{R}^3$ are derived via a deterministic coordinate transformation: $\boldsymbol{\mu}_g = \boldsymbol{R}_i \boldsymbol{P}_i + \boldsymbol{T}_i$. Parameterizing the scene geometry through dense point maps rather than depth maps inherently preserves the local metric structure and mitigates topological distortions~\cite{shi2025revisiting,wang2024dust3r,ramamonjisoa2020predicting,sun2023sc}, as visualized in~\cref{fig:anysplat_cmp}.

For the remaining rendering attributes, a Dense Prediction Transformer (DPT) head \(H_\text{G}\) is utilized~\cite{ranftl2021vision}. This head integrates the deep equivariant tokens \(F_{\text{DPT}}(\boldsymbol{h}_i)\) with high-resolution shallow features \(F_{\text{RGB}}(I_i)\) to regress opacity \(\sigma_g\), quaternion rotation \(\boldsymbol{r}_g\), anisotropic scales \(\boldsymbol{s}_g\), and spherical harmonic coefficients \(\boldsymbol{c}_g\). Formulated as:
\begin{equation}
\begin{cases}
    (\boldsymbol{P}_i, \boldsymbol{C}_i) = D_{\text{p}}(\boldsymbol{h}_i),\; \boldsymbol{\mu}_g = \text{trans}(\boldsymbol{P}_i, \boldsymbol{p}_i),  \\
    (\sigma_g, \boldsymbol{r}_g, \boldsymbol{s}_g, \boldsymbol{c}_g) = H_\text{G}(F_{\text{DPT}}(\boldsymbol{h}_i) + F_{\text{RGB}}(I_i)).
\end{cases}
\end{equation}

\subsection{Pose-Free Pre-training via 3D-Prior Distillation} 
Directly deploying feed-forward 3D Gaussian Splatting in the robotic domain is frequently hindered by severe ghosting and rendering artifacts. These issues stem from the inability of pure photometric optimization to preserve strict geometric consistency~\cite{ye2024no}, a critical prerequisite for reliable robotic manipulation. Compounding this challenge, existing large-scale robotic datasets are typically characterized by noisy or entirely missing camera pose annotations. 

To address this, we introduce a 3D-prior distillation strategy that extracts geometric knowledge from a pre-trained visual geometry foundation model~\cite{wang2025pi}, providing essential structural supervision for pose-free novel view synthesis. Specifically, to parameterize these spatial priors, we regress per-pixel 3D coordinates in the camera frame using pseudo-ground-truth from the foundation model, as detailed in~\cref{fig:pipeline}. The point map loss $\mathcal{L}_{\text{3D}}$ is then computed as the mean squared Euclidean distance between the predicted geometry $\hat{\boldsymbol{P}}$ and the target $\boldsymbol{P}$:
\begin{equation}
    \mathcal{L}_{\text{3D}} = \frac{1}{HW} \sum_{u,v} \left\| \hat{\boldsymbol{P}}(u,v) - \boldsymbol{P}(u,v) \right\|_2^2,
\end{equation}
where $(u,v)$ denotes the spatial pixel coordinates. As empirically visualized in~\cref{fig:2.2}, this explicit 3D supervision significantly reduces novel-view artifacts, such as floaters and boundary tearing, in Gaussian rendering. It markedly outperforms both depth-based 2.5D supervision~\cite{jiang2025anysplat,sun2025uni3r} and RGB-only baselines~\cite{fan2024instantsplat,ye2024no}. The resulting multi-view geometric coherence yields clean, low-variance novel-view training data. This metric stability is essential for reliably expanding the observational manifold and enhancing view-generalized execution for downstream policies.

To construct a consistent global coordinate frame without absolute tracking data, we distill camera-pose priors to obtain relative affine-invariant transformations. For the relative pose between views $i$ and $j$, rotational alignment is supervised via the geodesic distance $\ell_{\text{rot}}$ on the $SO(3)$ manifold~\cite{chowdhury2022unsupervised}, and translational alignment is constrained using a robust Huber loss $\mathcal{H}_{\delta}$ applied to the relative translation vectors:
\begin{equation}
\mathcal{L}_{\text{cam}} \!=\! \sum_{i \neq j} \Big( \ell_{\text{rot}}(\boldsymbol{\hat{R}}_{i\leftarrow j}, \boldsymbol{R}_{i\leftarrow j}) +  \mathcal{H}_{\delta}(\boldsymbol{\hat{T}}_{i\leftarrow j} - \boldsymbol{T}_{i\leftarrow j})\Big).
\label{eq:cam_loss}
\end{equation}

This affine-invariant pose supervision exploits the inherent low-dimensional manifold of real-world camera trajectories~\cite{wang2025pi}, regularizing geometry, suppressing drift and degeneracies, and stabilizing novel-view synthesis from sparse inputs. The complete loss \(\mathcal{L}_{\text{3D-prior}}\) is defined as the weighted sum of the point map loss and the camera pose loss:
\begin{equation}
\mathcal{L}_{\text{3D-prior}} = \lambda_{\text{3D}} \mathcal{L}_{\text{3D}} + \lambda_{\text{cam}} \mathcal{L}_{\text{cam}}.
\label{eq:prior_loss}
\end{equation}

Ultimately, the proposed distillation strategy yields substantial advantages for both scene reconstruction and policy learning. Fundamentally, point maps serve as a 2D parameterization of 3D features, coupling metric precision with structural regularity to provide an optimal supervisory signal. Unlike traditional depth-based pseudo-labels that exhibit sharp boundary discontinuities and cause tearing artifacts~\cite{jiang2025anysplat}, dense point maps explicitly enforce spatial smoothness and topological coherence. Distilling these priors imposes robust structural constraints, accelerating convergence and producing the high-fidelity representations essential for reliable robotic manipulation.

\subsection{Policy Deployment}
\label{Policy Deployment}
We instantiate our view-generalized framework across two representative imitation learning architectures: $\pi_0$~\cite{black2410pi0} and Diffusion Policy~\cite{chi2023diffusion}. Both models are optimized over the expanded observational manifold $\mathcal{D}_{\text{train}}$. Formally, the policy $\pi_\theta$ processes a visual observation $\boldsymbol{o}$ alongside the robot's proprioceptive state to predict a sequence of future kinematic actions $\boldsymbol{a}$. The learning objective follows a standard behavioral cloning paradigm, maximizing the log-likelihood of expert demonstrations under the parameterized policy:
\begin{equation}
    \max_{\theta}\sum_{(\boldsymbol{o},\,\boldsymbol{a}) \in \mathcal{D}_{\text{train}}} \log \pi_{\theta}(\boldsymbol{a}\mid \boldsymbol{o}),
\end{equation}
where $\theta$ denotes the learnable parameters of the policy network. Comprehensive training configurations and architectural details are deferred to Appendix \textcolor{red}{A.1}.

\subsection{Training Setup}
\label{Training Setup}
\textbf{Training Objectives.}
GenSplat is optimized end-to-end by minimizing a synergistic combination of photometric, geometric, and distillation objectives:
\begin{equation}
\mathcal{L} = \mathcal{L}_{\text{rgb}} + \lambda_{\text{depth}} \mathcal{L}_{\text{depth}} + \mathcal{L}_{\text{3D-prior}}.
\label{eq:total_loss}
\end{equation}

For each input view, the photometric rendering pipeline is strictly supervised via a combination of an $\ell_1$ penalty and a perceptual LPIPS regularizer~\cite{zhang2018unreasonable}:
\begin{equation}
\mathcal{L}_{\text{rgb}} = \lambda_{\text{1}}\| I - \hat{I} \|_1 + \lambda_{\text{2}} \, \mathtt{LPIPS}(I,\hat{I}),
\label{eq:rgb_loss}
\end{equation}
where $\hat{I}$ denotes the novel view synthesized via the differentiable splatting process.
To mitigate the subtle ambiguities that arise during multi-view alignment and aggregation~\cite{jiang2025anysplat}, we introduce a depth alignment loss that enforces alignment between the Gaussian primitives and the predicted point map, promoting geometric stability and mitigating view inconsistency artifacts:
\begin{equation}
\mathcal{L}_{\text{depth}}
= \frac{1}{\lVert M \rVert_{1}}
\left\lVert\, M \odot \big( \boldsymbol{P}_{...,\;2} - \hat{D} \big) \,\right\rVert_{2}^{2},
\end{equation}
where \(M\) is a binary mask indicating valid pixels, \(\hat{D}\) denotes the rendered depth, and the predicted depth is obtained from the z-component of the point map.

\textbf{Implementation Details.}
The aggregation transformer contains 36 alternating attention blocks that interleave frame-wise and global attention, and camera pose, point map decoders are lightweight 5-layer Transformer modules that use self-attention only. 
All these layers together with the prediction decoder introduced in~\cref{sec3.2 Model Architecture} are initialized from~\cite{wang2025pi} so that the model inherits permutation equivariant geometric priors and attains stable early training. 
Moreover, to narrow the gap between general scenes and robot manipulation, we pretrain on a large-scale dataset collected in the wild~\cite{khazatsky2024droid}. We temporally subsample every episode at 1 frame per second, which yields 271k training images that cover a wide range of environments and tasks. 
We train for 30k iterations with a batch size of 16 and maintain an exponential moving average of the model weights to stabilize evaluation. 
The learning rate follows a cosine annealing schedule with a 1k step linear warmup, and the peak learning rate equals \(2e^{-4}\).
Following~\cite{jiang2025anysplat}, parameters initialized from ~\cite{wang2025pi} use a learning rate of 0.1 times the base value, and we enable FlashAttention~\cite{dao2022flashattention}, gradient checkpointing, and bfloat16 precision to reduce memory usage and accelerate training.

\section{Experiments}
\label{sec:Experiments}

We conduct extensive experiments on a real-world robotic platform to evaluate the performance of our proposed method, GenSplat. Our evaluation is structured around four central research questions (RQs) designed to assess our contributions systematically:

\begin{enumerate}[label=RQ\arabic*:, leftmargin=*]
    \item How effectively does expanding the observational manifold via GenSplat improve policy robustness to out-of-distribution camera perturbations?
    \item How does GenSplat compare against diffusion-based and gaussian-based NVS methods in terms of rendering quality, policy learning, and efficiency?
    \item How does the core 3D-prior distillation component impact geometric consistency, rendering fidelity, and the success rate of downstream policies?
    \item What is the impact of increasing the number of rendered views per trajectory on view generalization and data efficiency?
\end{enumerate}

\begin{figure*}[t]
    \centering
    \includegraphics[width=\linewidth]{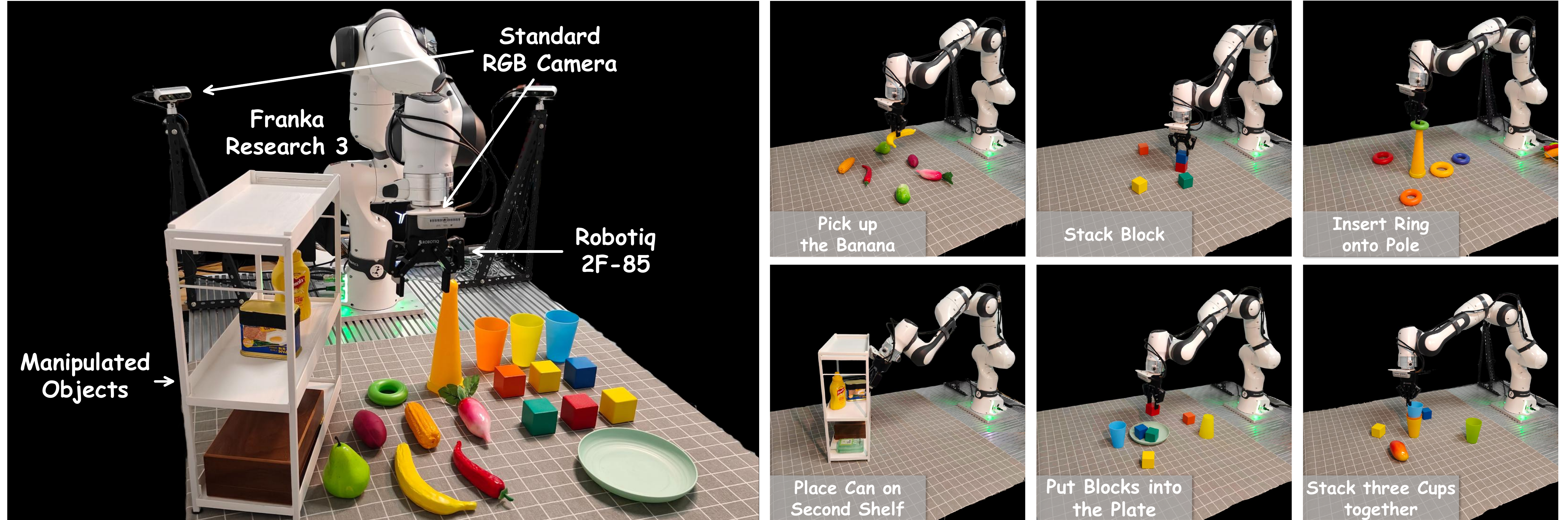}
    \caption{\textbf{Overview of the experiment setup and tasks.} We design six manipulation tasks for real-world evaluation. 
    }
    \label{fig:real}
    \vspace{-2ex}
\end{figure*}

\noindent \textbf{Experimental Setup.} 
Our hardware suite (\cref{fig:real}) comprises a 7-DoF Franka Research 3 arm equipped with a Robotiq 2F-85 gripper. Visual observations are captured by three standard RGB cameras: two providing external views and one mounted on the robot's wrist. 
We evaluate our framework across six manipulation tasks. For each task, we collect 100 expert demonstrations via teleoperation. To ensure rigorous evaluation, the dataset incorporates rich spatial and visual diversity through systematic variations in object positions and colors.

\subsection{RQ1: Robustness to Camera Perturbations}
\label{sec:Robustness to Camera Perturbations}
This experiment quantitatively investigates whether densifying the spatial support via GenSplat systematically improves policy robustness against camera perturbations of varying magnitudes.

\noindent \textbf{Viewpoint re-rendering strategy.}
We employ GenSplat to reconstruct the scene and synthesize novel-view trajectories while strictly preserving temporal alignment and kinematic action labels. For each demonstration, we define the pivot point as the midpoint of the shortest line segment connecting the optical axes of the two external cameras. 
Following~\cite{tian2024view,shi2025nvspolicy}, novel views are generated by rotating the camera about a designated world axis ($\mathcal{X}$ or $\mathcal{Y}$) by $\theta \in [-30^\circ, 30^\circ]$, followed by a forward translation $\mu \in [0.5, 6]$ cm along the camera's $+\mathcal{Z}$ axis. This translation crucially prevents severe rotations from pushing the viewpoint outside the densified spatial support (\textit{i.e.}, beyond valid splat coverage). For each original demonstration, we sample exactly one unique perturbation pair $(\theta,\mu)$ to generate a single augmented trajectory.

\noindent \textbf{Perturbation analysis protocol.}
We deploy the trained policies under three controlled spatial perturbation regimes during physical rollouts: 
(1) \textbf{Small}: Rotation $\theta_{\max}\in[3^\circ,6^\circ]$ and translation $\mu\in[0.5,2]$ cm. 
(2) \textbf{Medium}: Rotation $\theta_{\max}\in[8^\circ,15^\circ]$ and translation $\mu\in[2,4]$ cm. 
(3) \textbf{Large}: Rotation $\theta_{\max}\in[18^\circ,30^\circ]$ and translation $\mu\in[4,6]$ cm. We compare baselines trained strictly on source data against our policies trained on the expanded manifold.

\begin{figure*}[t]
    \centering
    \includegraphics[width=\linewidth]{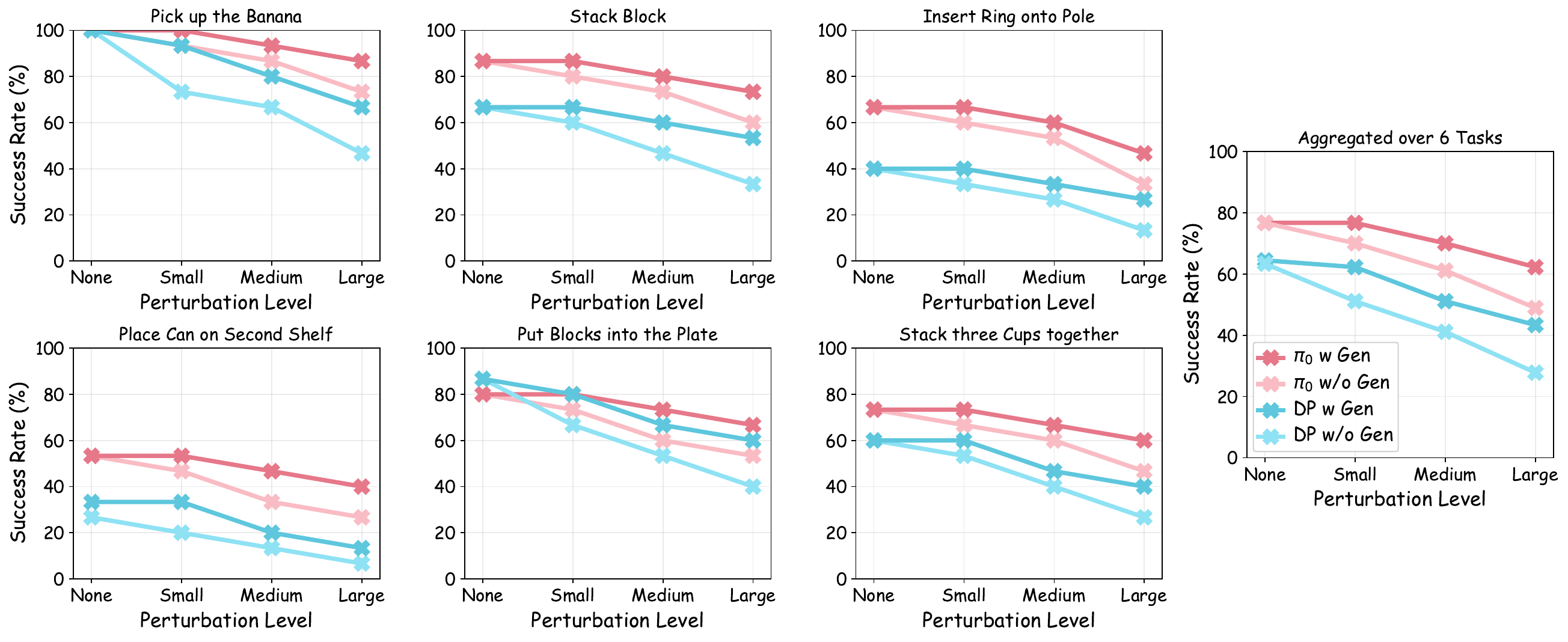}
        \caption{\textbf{Policy robustness to camera pose perturbations.} We evaluate 30 episodes per task on 6 real-world tasks under increasing camera perturbations and report mean success rates (in \%). Policies trained with GenSplat augmented novel views data consistently outperform those trained only on source views for both \(\pi_{0}\) and Diffusion Policy. 
    }
    \vspace{-2ex}
    \label{fig:Q2}
\end{figure*}

As visualized in~\cref{fig:Q2}, GenSplat augmentation yields consistent and substantial improvements in policy robustness across all six manipulation tasks and all perturbation levels, mitigating the sharp degradation typically caused by out-of-distribution camera viewpoints. Aggregated across all tasks, GenSplat yields substantial relative performance improvements across all perturbation levels for \(\pi_0\): +\textbf{6.7}\% for small, +\textbf{8.9}\% for medium, and +\textbf{13.3}\% for large perturbations. The benefit is most pronounced for DP under large perturbations, where source-view-only policies suffer significant collapse, and GenSplat achieves a remarkable \textbf{56.0}\% relative gain (from 27.78\% to \textbf{43.33}\%). Crucially, the augmented policies maintain parity with the baselines in the unperturbed (source) setting, confirming that our geometrically consistent novel views do not compromise the original data distribution. These results establish GenSplat as an effective, architecture-agnostic strategy for training viewpoint generalizable policies.

\vspace{-2ex}
\subsection{RQ2: Comprehensive Comparison with SOTA}
We address RQ2 by evaluating GenSplat across three critical axes: (1) rendering quality, (2) downstream policy performance, and (3) synthesis efficiency. We compare GenSplat against leading NVS methods including ZeroNVS~\cite{sargent2024zeronvs}, VISTA~\cite{tian2024view}, SEVA~\cite{zhou2025stable}, InstantSplat~\cite{fan2024instantsplat}, NoPoSplat~\cite{ye2024no} and AnySplat~\cite{jiang2025anysplat}.

\begin{figure}[t]
    \centering
    \includegraphics[width=1.0\linewidth]{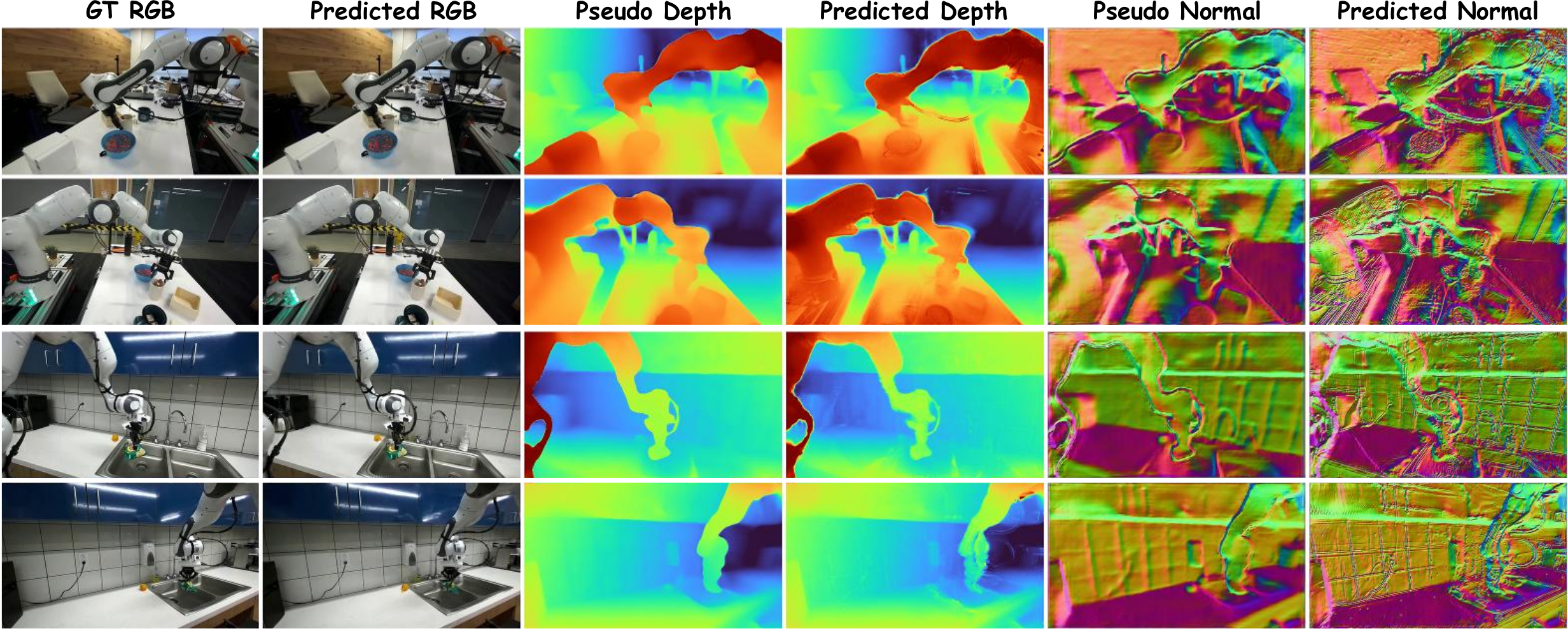}
    \caption{\textbf{Qualitative 3D geometry reconstruction on the DROID dataset.} We visualize pairs of reference targets (ground truth or pseudo-labels) against GenSplat's predicted RGB, depth, and normal maps. GenSplat accurately recovers structural high-frequency details and maintains strict spatial alignment without explicit camera calibration. 
    Extended visual comparisons are provided in Appendix \textcolor{red}{B}.}
    \vspace{-3ex}
    \label{fig:2.1}
\end{figure}

\noindent \textbf{Geometric Consistency \& Rendering Quality.} 
We first validate the underlying 3D structural fidelity. As illustrated in~\cref{fig:2.1}, GenSplat accurately infers dense depth and normal maps. While the distilled pseudo-labels inherently suffer from over-smoothing, our unified photometric and geometric optimization effectively recovers sharp object boundaries and fine-grained topologies. By leveraging RGB photometric supervision with 3D-priors strictly acting as structural regularizers, GenSplat successfully reconstructs high-frequency details.

\begin{figure}[t]
    \centering
    \includegraphics[width=1.0\linewidth]{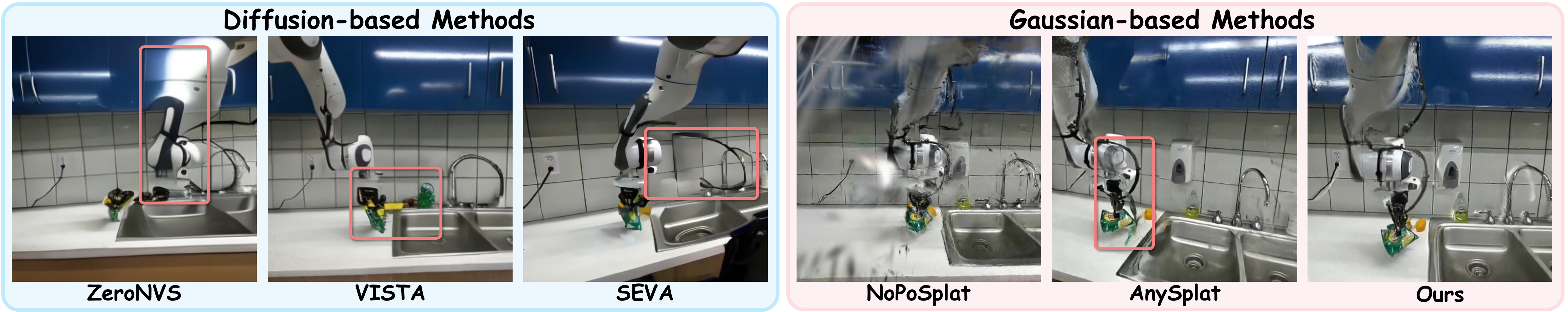}
    \caption{\textbf{Qualitative comparison of NVS.} 
    We compared GenSplat with diffusion- and Gaussian-based methods under the same input example. GenSplat shows a clear advantage in 3D geometric consistency modeling while effectively eliminating floating artifacts and visual artifacts. Note: VISTA is built on ZeroNVS and is fine-tuned on a 3k trajectory subset of DROID~\cite{khazatsky2024droid}, yielding significant improvements in NVS.
    }
    \vspace{-4ex}
    \label{fig:2.2}
\end{figure}

Building upon this robust geometry, GenSplat demonstrates a significant advantage in synthesizing high-fidelity novel views. As depicted in~\cref{fig:2.2}, diffusion-based baselines frequently violate strict epipolar geometry, resulting in severe spatial hallucinations and metric distortions (\textit{e.g.}, the bent faucet). Conversely, conventional Gaussian-based methods lacking prior distillation are plagued by severe floater artifacts and boundary tearing~\cite{ye2024no}, fundamentally degrading visual quality. In stark contrast, GenSplat renders scenes with photorealism and absolute structural integrity. This geometrically consistent, high-fidelity synthesis is paramount, providing the clean, low-variance multi-view data necessary for reliable downstream policy learning.


\definecolor{lightgreen}{HTML}{EDF8FE}
\definecolor{grey}{RGB}{235,235,235}

\begin{table*}[t]
    \centering
    \resizebox{\linewidth}{!}{
        \setlength{\tabcolsep}{0.5em}
        \begin{tabular}{l | c | c c c c c c | c}
        \toprule
        \rowcolor{lightgreen}
        \multicolumn{1}{l|}{Method / Task} & 
        \multicolumn{1}{c|}{\begin{tabular}[c]{@{}c@{}}Avg.\\ Success↑\end{tabular}} & 
        \multicolumn{1}{c}{\begin{tabular}[c]{@{}c@{}}Pick up\\ Banana\end{tabular}} & 
        \multicolumn{1}{c}{\begin{tabular}[c]{@{}c@{}}Stack\\ Block\end{tabular}} & 
        \multicolumn{1}{c}{\begin{tabular}[c]{@{}c@{}}Insert\\ Ring\end{tabular}} & 
        \multicolumn{1}{c}{\begin{tabular}[c]{@{}c@{}}Place \\Can \end{tabular}} & 
        \multicolumn{1}{c}{\begin{tabular}[c]{@{}c@{}}Put Blocks\\ into Plate\end{tabular}} & 
        \multicolumn{1}{c|}{\begin{tabular}[c]{@{}c@{}}Stack\\ Cups\end{tabular}} & 
        \multicolumn{1}{c}{\begin{tabular}[c]{@{}c@{}}Synthesis\\ Speed\end{tabular}} \\
        \midrule

        \rowcolor{grey}
        Baseline & 48.9 & 73.3 & 60.0 & 33.3 & 26.7 & 53.3 & 46.7 & - \\

        VISTA~\cite{tian2024view} & 37.8 & 76.7 & 43.3 & 36.7 & 6.7 & 40.0 & 23.3 & 2.81 s \\

        SEVA~\cite{zhou2025stable} & 57.7 & 83.3 & 70.0 & 43.3 & 30.0 & 63.3 & 56.7 & 50.00 s \\

        InstantSplat~\cite{fan2024instantsplat} & 55.0 & 80.0 & 66.7 & 40.0 & 33.3 & 60.0 & 50.0 & 85.20 s \\

        NoPoSplat~\cite{ye2024no} & 36.7 & 73.3 & 40.0 & 33.3 & 10.0 & 36.7 & 26.7 & \textbf{0.05} s \\

        AnySplat~\cite{jiang2025anysplat} & 56.1 & 80.0 & 66.7 & 43.3 & 30.0 & 63.3 & 53.3 & 0.16 s \\

        \midrule 
        GenSplat(ours) & \textbf{62.2} & \textbf{86.7} & \textbf{73.3} & \textbf{46.7} & \textbf{40.0} & \textbf{66.7} & \textbf{60.0} & 0.14 s \\ 
        \bottomrule 

        \end{tabular}
    }
    \caption{\textbf{Policy Performance and Synthesis Efficiency.} We compare various NVS methods by training \(\pi_0\) on augmented datasets and report mean success rate and synthesis speed. Each task was evaluated 30 episodes under the \textbf{Large} perturbation level. Synthesis speed evaluated on the same computing device.}
    \label{tab:RQ4}
    \vspace{-6ex}
\end{table*}

\noindent \textbf{Downstream Policy Performance.} 
As detailed in~\cref{tab:RQ4}, GenSplat achieves the highest average success rate of \textbf{62.2}\% under large camera perturbations, significantly outperforming all competing NVS paradigms. Crucially, the results reveal that low-quality data augmentation actively harms policy learning: both the diffusion-based VISTA~\cite{tian2024view} (37.8\%) and the standard feed-forward NoPoSplat~\cite{ye2024no} (36.7\%) severely degrade performance below the unaugmented Baseline (48.9\%). Their pronounced geometric distortions and rendering artifacts act as noise, contaminating the visuomotor mapping (see~\cref{fig:2.2}). While SEVA~\cite{zhou2025stable} (57.7\%) and InstantSplat~\cite{fan2024instantsplat} (55.0\%) offer improvements, their sparse-view reconstructions still exhibit localized holes and inconsistencies that bottleneck execution accuracy. Compared to the closest feed-forward baseline, AnySplat~\cite{jiang2025anysplat} (56.1\%), GenSplat provides a substantial absolute gain of \textbf{+6.1}\%. This confirms that our 3D-prior distillation explicitly enforces the spatial smoothness necessary to generate physically reliable, high-fidelity training signals.

\noindent \textbf{Synthesis Efficiency.} 
For scalable data augmentation in robotics, computational efficiency is a hard constraint. GenSplat operates at a highly efficient 0.14 seconds per frame, completely bypassing the prohibitive computational bottlenecks of iterative diffusion sampling and per-scene optimization. While NoPoSplat~\cite{ye2024no} is marginally faster, this speed incurs the unacceptable cost of catastrophic policy degradation. 
Ultimately, GenSplat achieves remarkable generation speed without sacrificing geometric consistency, positioning our framework as a highly scalable engine for learning view-generalized robotic policies.

\vspace{-2ex}
\subsection{RQ3: Ablation of 3D-Prior Distillation}
To address RQ3, we conduct an ablation study isolating the contributions of the permutation-equivariant (P.E.) architecture \cite{wang2025pi} and the core 3D-prior distillation module. \cref{tab:ablation} details the quantitative impact of these components on both novel view synthesis fidelity and downstream policy robustness under challenging viewpoint shifts.
Replacing the baseline VGGT with the P.E. architecture improves reconstruction. By mitigating order-dependent artifacts from uncalibrated inputs, it boosts policy success from 66.0\% to 70.0\% under large perturbations, and 30.0\% to 40.0\% under extreme shifts. 
Furthermore, integrating 3D-prior distillation provides essential geometric regularization for pose-free settings, achieving the best PSNR and LPIPS. This strict structural consistency translates directly to physical robustness, maximizing policy success to \textbf{74.0\%} (large) and \textbf{46.0\%} (extreme). By explicitly suppressing floating artifacts, the 3D prior prevents geometric collapse, delivering the high-fidelity manifold data necessary to induce view-generalized behaviors.
\renewcommand{\arraystretch}{1.0} 

\definecolor{darkgreen}{HTML}{25C445}
\definecolor{darkred}{HTML}{DC143C}
\definecolor{lightgrey}{HTML}{E8E8E8}
\definecolor{final}{HTML}{62EF7E}
\definecolor{base}{HTML}{BDBDBD}

\begin{table}[t]
    \centering
    \resizebox{0.9\linewidth}{!}{
        \setlength{\tabcolsep}{1.0em} 
        \begin{tabular}{ccc|cccc} 
        \toprule
        \rowcolor{lightgreen}
        VGGT & P.E.~\cite{wang2025pi} & \begin{tabular}[c]{@{}c@{}} 3D-Prior \end{tabular} & \begin{tabular}[c]{@{}c@{}} PSNR↑ \end{tabular} & \begin{tabular}[c]{@{}c@{}}LPIPS↓\end{tabular} & \begin{tabular}[c]{@{}c@{}} $30^{\circ}$, 6cm \end{tabular} & \begin{tabular}[c]{@{}c@{}} $60^{\circ}$, 12cm \end{tabular} \\ 
        \midrule
         
        {\color{darkgreen} \Checkmark} & {\color{darkred} \XSolidBrush} & {\color{darkred} \XSolidBrush} & 22.35 & 6.01 & 66.0\% & 30.0\% \\
         
        {\color{darkred} \XSolidBrush} & {\color{darkgreen} \Checkmark} & {\color{darkred} \XSolidBrush} & 25.87 & 5.03 & 70.0\% & 40.0\% \\
        
        {\color{darkred} \XSolidBrush} & {\color{darkgreen} \Checkmark} & {\color{darkgreen} \Checkmark} & \textbf{26.53} & \textbf{3.72} & \textbf{74.0\%} & \textbf{46.0\%} \\
         
         \bottomrule
        \end{tabular}
    }
    \caption{\textbf{Ablation of GenSplat components.} We evaluate the impact of the permutation-equivariant (P.E.) architecture and 3D-prior distillation on novel view rendering quality (PSNR, LPIPS) and policy average success rates under \textbf{Large} ($30^\circ$, 6cm) and \textbf{Extreme} ($60^\circ$, 12cm) camera perturbations.}
    \label{tab:ablation}
    \vspace{-7ex}
\end{table}

\vspace{-4.5ex}

\subsection{RQ4: Impact of the Number of Rendered Views}
\begin{figure*}[h!]
    \centering
    \vspace{-3ex}
    \includegraphics[width=0.95\linewidth]{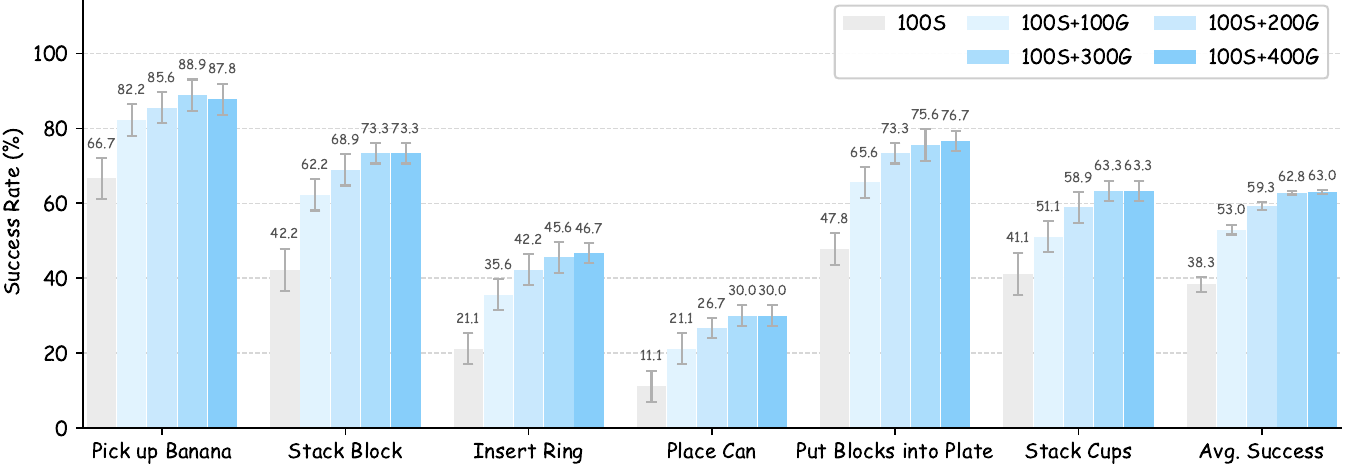}
    \vspace{-3ex}
    \caption{\textbf{Effect of Rendered Views on Viewpoint Robustness.}
    We report the mean success rate and standard deviation across three independent trials, with 30 execution episodes per task in each run. 
    }
    \vspace{-3ex}
    \label{fig:Q3}
\end{figure*}

For RQ4, we quantify the trade-off between the number of synthesized novel views and policy generalization. Fixing the source trajectories at 100, we systematically vary the view augmentation factor \(N \in \{1, 2, 3, 4\}\) and evaluate the resulting Diffusion Policies~\cite{chi2023diffusion}. According to~\cref{fig:Q3}, under a \textbf{Medium} perturbation regime, increasing the number of rendered views consistently improves policy success rates. Notably, the most substantial performance leap occurs immediately upon introducing the first augmented view ($N=1$).

\begin{table}[h]
    \centering
    \begin{tabularx}{0.9\linewidth}{ l *{8}{>{\centering\arraybackslash}X} }
        \toprule
        \rowcolor{lightblue}
        Task & \multicolumn{4}{c}{Stack Block} & \multicolumn{4}{c}{Stack Cups} \\
        \rowcolor{lightblue}
        Perturbation & None & Small & Medium & Large & None & Small & Medium & Large \\
        \midrule
        
        \rowcolor{grey}
        100S (Baseline) & 68.0 & 58.0 & 42.0 & 30.0 & 60.0 & 50.0 & 40.0 & 26.0 \\
        100S+100G       & 70.0 & 66.0 & 62.0 & 48.0 & 62.0 & 58.0 & 50.0 & 44.0 \\
        100S+200G       & 68.0 & 70.0 & 68.0 & 58.0 & 60.0 & 62.0 & 58.0 & 50.0 \\
        100S+300G       & \textbf{72.0} & \textbf{74.0} & \textbf{74.0} & \textbf{64.0} & \textbf{64.0} & 64.0 & \textbf{64.0} & \textbf{56.0} \\
        100S+400G       & 70.0 & 74.0 & 72.0 & 62.0 & 62.0 & \textbf{66.0} & 64.0 & 54.0 \\
        
        \bottomrule
    \end{tabularx}
    \caption{\textbf{Cross-Perturbation Analysis.} We report the success rates for Diffusion Policy with varying GenSplat view augmentations, evaluated on two tasks across perturbation levels from None to Large, based on 50 episodes per task.}
    \label{tab:RQ3-2}
    \vspace{-5ex}
\end{table}

\begin{figure*}[ht!]
    \centering
    \includegraphics[width=0.9\linewidth]{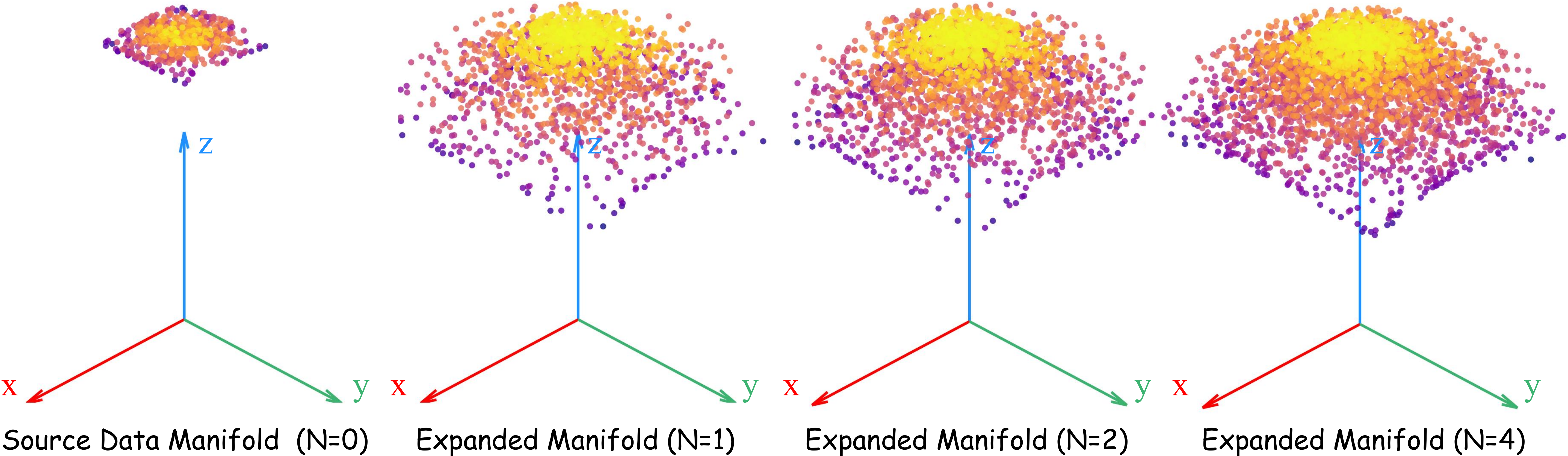}
    \caption{
    \textbf{Densification of camera extrinsics on the observational manifold.}
    }
    \vspace{-4ex}
    \label{fig:fig_manifold}
\end{figure*}

\noindent \textbf{Analysis of Data Efficiency.}
Critically, performance gains saturate rapidly. \cref{tab:RQ3-2} extends this analysis across all perturbation severities on the \texttt{Stack Block} and \texttt{Stack Cups} tasks, revealing a consistent scaling law: after a sharp initial leap at $N=1$, marginal benefits diminish and saturate near $N=3$. Since $N=4$ yields negligible improvements, a moderate augmentation density ($N=3$) is proven sufficient to unlock robust view generalization, effectively circumventing the overhead of excessive rendering.

\noindent \textbf{Geometric Interpretation.}
To geometrically explain this saturation, \cref{fig:fig_manifold} visualizes a single camera's extrinsics across 600 real-world demonstrations. The initial augmentation ($N=1$) sharply expands the sparse source distribution ($N=0$), driving the primary performance leap. Subsequent views ($N \in \{2, 4\}$) merely densify this expanded spherical manifold without enlarging its boundary. Once topological continuity is achieved (near $N=3$), the policy interpolates robustly, rendering further augmentations computationally redundant.

\section{Conclusion and Discussion}
\label{sec:Conclusion}
\vspace{-1ex}


We introduced GenSplat, a feed-forward 3D Gaussian Splatting framework synthesizing geometrically consistent novel views from sparse, uncalibrated observations. 
By explicitly incorporating 3D-prior distillation, GenSplat eliminates the structural artifacts inherent to pure photometric optimization. 
Driven by its feed-forward design, this high-fidelity and efficient synthesis provides a scalable data augmentation paradigm, expanding the observational manifold to resolve the critical bottleneck of policy view generalization. 
Future work will explore integrating dynamic wrist-camera observations to capture fine-grained, occlusion-aware geometries during active manipulation.


\clearpage

%
%
\bibliographystyle{splncs04}
\bibliography{ref}

\end{document}